\documentclass{article}
\usepackage{ijcai25}

\usepackage{times}
\usepackage{soul}
\usepackage{url}
\usepackage[hidelinks]{hyperref}
\usepackage[utf8]{inputenc}
\usepackage[small]{caption}
\usepackage{graphicx}
\usepackage{amsmath}
\usepackage{amsthm}
\usepackage{booktabs}
\usepackage{algorithm}
\usepackage{algorithmic}
\usepackage[switch]{lineno}
\usepackage{amsfonts}
\usepackage{natbib}

\usepackage{pifont}
\usepackage{tikz}
\usepackage[edges]{forest}
\usepackage{multirow}
\usepackage{booktabs}
\definecolor{hidden-draw}{RGB}{106,142,189} 
\definecolor{hidden-blue}{RGB}{194,232,247} 
\definecolor{hidden-orange}{RGB}{217, 232, 252} 
\newcommand{\ie}{\mbox{\it{i.e.,\ }}}
\newcommand{\eg}{\mbox{\it{e.g.,\ }}}

\def\Snospace~{\S{}}

\newcommand{\sref}[2]{\hyperref[#2]{#1 \ref{#2}}}

\usepackage{xcolor}
\definecolor{yucky}{HTML}{a64d79}
\newcommand*\circled[1]{\scalebox{0.8}{\tikz[baseline=(char.base)]{
\node[anchor=text, shape=circle,fill=yucky, inner sep=0pt, minimum size=1.2em] (char) {\footnotesize \textcolor{white}{#1}};}}}
\definecolor{darkgreen}{rgb}{0,0.40,0}
\definecolor{firebrick}{rgb}{0.698,0.133,0.133}
\hypersetup{
    colorlinks=true,
    linkcolor=firebrick,
    citecolor=darkgreen,
    filecolor=firebrick,      
    urlcolor=firebrick,
}

\title{A Survey on Explainable Deep Reinforcement Learning}

\author{
Zelei Cheng$^1$\thanks{These authors contributed equally to this work.}
\and
Jiahao Yu$^1$\footnotemark[1] \and
Xinyu Xing$^1$\\
\affiliations
$^1$Northwestern University\\
\emails
\{zelei.cheng, jiahao.yu, xinyu.xing\}@northwestern.edu
}

\begin{document}

\maketitle

\begin{abstract}
Deep Reinforcement Learning (DRL) has achieved remarkable success in sequential decision-making tasks across diverse domains, yet its reliance on black-box neural architectures hinders interpretability, trust, and deployment in high-stakes applications. Explainable Deep Reinforcement Learning (XRL) addresses these challenges by enhancing transparency through feature-level, state-level, dataset-level, and model-level explanation techniques. This survey provides a comprehensive review of XRL methods, evaluates their qualitative and quantitative assessment frameworks, and explores their role in policy refinement, adversarial robustness, and security. Additionally, we examine the integration of reinforcement learning with Large Language Models (LLMs), particularly through Reinforcement Learning from Human Feedback (RLHF), which optimizes AI alignment with human preferences. We conclude by highlighting open research challenges and future directions to advance the development of interpretable, reliable, and accountable DRL systems.
\end{abstract}
\section{Introduction}

Deep Reinforcement Learning (DRL) has emerged as a transformative paradigm for solving complex sequential decision-making problems. By enabling autonomous agents to interact with an environment, receive feedback in the form of rewards, and iteratively refine their policies, DRL has demonstrated remarkable success across a diverse range of domains including games (\eg Atari~\citep{mnih2013playing,kaiser2020model}, Go~\citep{silver2018general,silver2017mastering}, and StarCraft II~\citep{vinyals2019grandmaster,vinyals2017starcraft}), robotics~\citep{kalashnikov2018scalable}, communication networks~\citep{feriani2021single}, and finance~\citep{liu2024dynamic}. These successes underscore DRL's capability to surpass traditional rule-based systems, particularly in high-dimensional and dynamically evolving environments.

Despite these advances, a fundamental challenge remains: DRL agents typically rely on deep neural networks, which operate as black-box models, obscuring the rationale behind their decision-making processes. This opacity poses significant barriers to adoption in safety-critical and high-stakes applications, where interpretability is crucial for trust, compliance, and debugging. The lack of transparency in DRL can lead to unreliable decision-making, rendering it unsuitable for domains where explainability is a prerequisite, such as healthcare, autonomous driving, and financial risk assessment.

To address these concerns, the field of Explainable Deep Reinforcement Learning (XRL) has emerged, aiming to develop techniques that enhance the interpretability of DRL policies. XRL seeks to provide insights into an agent’s decision-making process, enabling researchers, practitioners, and end-users to understand, validate, and refine learned policies. By facilitating greater transparency, XRL contributes to the development of safer, more robust, and ethically aligned AI systems.

Furthermore, the increasing integration of Reinforcement Learning (RL) with Large Language Models (LLMs) has placed RL at the forefront of natural language processing (NLP) advancements. Methods such as Reinforcement Learning from Human Feedback (RLHF)~\citep{bai2022training,ouyang2022training} have become essential for aligning LLM outputs with human preferences and ethical guidelines. By treating language generation as a sequential decision-making process, RL-based fine-tuning enables LLMs to optimize for attributes such as factual accuracy, coherence, and user satisfaction, surpassing conventional supervised learning techniques. However, the application of RL in LLM alignment further amplifies the explainability challenge, as the complex interactions between RL updates and neural representations remain poorly understood.

This survey provides a systematic review of explainability methods in DRL, with a particular focus on their integration with LLMs and human-in-the-loop systems. We first introduce fundamental RL concepts and highlight key advances in DRL. We then categorize and analyze existing explanation techniques, encompassing feature-level, state-level, dataset-level, and model-level approaches. Additionally, we discuss methods for evaluating XRL techniques, considering both qualitative and quantitative assessment criteria. Finally, we explore real-world applications of XRL, including policy refinement, adversarial attack mitigation, and emerging challenges in ensuring interpretability in modern AI systems. Through this survey, we aim to provide a comprehensive perspective on the current state of XRL and outline future research directions to advance the development of interpretable and trustworthy DRL models.
\section{Preliminaries}
\label{sec:preliminaries}
\subsection{Reinforcement Learning Foundations}
Reinforcement Learning (RL) is a subfield of machine learning that focuses on training agents to make sequential decisions by interacting with an environment. The environment is framed as a Markov Decision Process (MDP) \citep{sutton2018reinforcement}, specified by the tuple \((\mathcal{S}, \mathcal{A}, P, \rho, R, \gamma)\):
\begin{itemize}
    \item \(\mathcal{S}\): A set of states representing possible configurations of the environment.
    \item \(\mathcal{A}\): A set of actions available to the agent.
    \item \(P(s' \mid s,a)\): The transition probability function describing how actions lead from one state \(s\) to another state \(s'\).
    \item $\rho$: the distribution of the initial state $s_0$.
    \item \(R(s,a)\): The immediate reward obtained after executing action \(a\) in state \(s\).
    \item \(\gamma \in (0,1)\): A discount factor that balances immediate and future rewards.
\end{itemize}

The goal of RL is to find an optimal policy $\pi(a|s)$: ($\mathcal{S} \rightarrow \mathcal{A}$) which maximizes the agent's long-term reward. Formally, the long-term reward is defined as the \emph{state-value function}
\begin{equation}
\small
V^{\pi}(s) =\sum_{a\in\mathcal{A}} \pi(a|s)\left[R(s,a)+\gamma \sum_{s'\in\mathcal{S}} P(s'|s,a)V^{\pi}(s')\right] \, .
\label{eq:state}
\end{equation}
Accordingly, the \emph{action-value function} $Q^{\pi}(s, a)$ is defined as
\begin{equation}
    \small
    Q^{{\pi}}(s, a) = R(s,a)+\gamma \sum_{s'\in\mathcal{S}} P(s'|s,a) \sum_{a'\in\mathcal{A}}\pi(a'|s')Q^{\pi}(s', a') \, .
    \label{eq:action}
\end{equation}

The \emph{advantage function} $A^{\pi} (s,a)$ is defined as
\begin{equation}
    \small
    A^{{\pi}}(s, a) = Q^{\pi}(s,a)-V^{\pi}(s) \, .
    \label{eq:advantage}
\end{equation}

In reinforcement learning, the state-value function $V^{\pi}(s)$ represents the expected total reward for an agent starting from state $s$. Slightly different from $V^{\pi}(s)$, the action-value function $Q^{\pi}(s, a)$ is the expected total reward for an agent to choose action $a$ while in state $s$. The advantage function measures the expected additional reward for choosing action $a$ over the expected reward of the policy.
The expected total reward of a policy $\pi$ is defined as 
\begin{equation}
    \small
    \eta(\pi)=\mathbb{E}_{s_0, a_0, \ldots}\left[\sum_{t=0}^{\infty} \gamma^t R\left(s_t, a_t\right)\right] \, .
    \label{eq:expected_reward}
\end{equation}
By maximizing the expected total reward, an optimal policy $\pi^{*}$ can be derived, enabling the agent to receive the maximum rewards in the environment.


Reinforcement learning can be categorized into two primary settings based on the agent's ability to interact with the environment: online RL and offline RL. In online RL, the agent has direct, interactive access to the environment and can continuously collect new experiences by executing and updating its policy in real-time. This setting allows for active exploration and immediate policy adaptation. In contrast, offline RL restricts the agent to learn solely from a fixed dataset of previously collected experiences, without any further environment interaction. This dataset typically consists of state-action-reward trajectories collected by one or multiple behavior policies. The offline setting is particularly relevant in scenarios where environment interaction is expensive, risky, or impractical, such as in healthcare, autonomous driving, or industrial control systems.

There are two types of main-stream algorithms, \ie value-based methods and policy-based methods. For value-based methods such as Q-learning algorithm~\citep{watkins1992q}, the agent estimates $Q(s,a)$ and greedily chooses the optimal action. Regarding policy-based methods, the agent directly optimizes its policy based on the reward feedback (\eg Policy Gradient methods~\citep{sutton1999policy}). Classic algorithms have been effective in relatively small or structured environments. However, their performance may degrade in high-dimensional or unstructured domains due to challenges in representation and exploration.

\subsection{Deep Reinforcement Learning Advancements}

To address the limitations of standard RL in complex or high-dimensional settings, Deep Reinforcement Learning (DRL) integrates neural networks as function approximators for policies or value functions. Two prominent approaches for learning deep reinforcement learning policies are Deep Q-Network (DQN)~\citep{mnih2015human} and Proximal Policy Optimization (PPO)~\citep{schulman2017proximal}. We provide a brief overview of the foundational principles underlying each of these algorithms.

\textbf{Deep Q-Network (DQN). }
DQN utilizes a deep neural network to approximate the optimal action-value function (Q function). The network architecture typically processes state inputs $s$ through several layers and outputs Q-values for all possible actions simultaneously. The network is trained by minimizing the temporal difference error between predicted and target Q-values using experience replay and a target network to stabilize training. During execution, the optimal policy is derived by selecting the action with the highest predicted Q-value.

\textbf{Proximal Policy Optimization (PPO).}
Different from DQN, policy gradient methods directly learn a parameterized policy $\pi_{\theta}(a|s) = \mathbb{P}(a|s,\theta)$ to maximize the expected total reward. While these methods offer more direct policy optimization, they often suffer from high variance and sensitivity to learning rates, leading to unstable training. PPO addresses these challenges by introducing a clipped surrogate objective function. It constrains policy updates to prevent excessive changes while optimizing performance. By maintaining proximity between consecutive policies and using advantage estimation, PPO achieves more stable training and better sample efficiency compared to traditional policy gradient methods, making it one of the most widely adopted algorithms in practice.



\subsection{Reinforcement Learning for LLMs}

The integration of RL with Large Language Models (LLMs) has emerged as a promising direction for improving the alignment and performance of AI systems. Multiple RL approaches such as PPO, Directed Preference Optimization (DPO)~\citep{rafailov2024direct}, Reward rAnked FineTuning (RAFT)~\citep{dong2023raft} have been used to fine-tune LLMs for specific tasks, such as dialogue generation, summarization, and instruction following. By leveraging reward feedback, RL-based approaches enable LLMs to generate more coherent, contextually appropriate, and user-aligned outputs.

Despite these advancements, the explainability of RL for LLMs remains an open challenge. The complexity of LLMs, combined with the sequential decision-making nature of RL, makes it difficult to interpret how the input data impacts these models to generate outputs. Recent efforts have explored techniques such as data influence functions to enhance the transparency of RL for LLMs. However, there is still a need for more systematic explanation approaches in this domain, particularly for applications involving ethical considerations, bias mitigation, and user trust.

In the subsequent sections, we survey existing methods for providing interpretability in DRL systems as well as LLMs, and discuss how these techniques can be evaluated and applied in practice.

\section{Explanation Techniques for DRL}

\tikzstyle{my-box}=[
 rectangle,
 draw=hidden-draw,
 rounded corners,
 text opacity=1,
 minimum height=1.5em,
 inner sep=2pt,
 align=center,
 fill opacity=.5,
]
\tikzstyle{leaf}=[my-box, 
 minimum height=1.5em,
 fill=hidden-orange!60, 
 text=black, 
 align=left,
 font=\scriptsize,
 inner xsep=2pt,
 inner ysep=4pt,
]

\begin{figure*}[t]
    \centering
    \resizebox{1\textwidth}{!}{
        \begin{forest}
            forked edges,
            for tree={
                grow=east,
                reversed=true,
                anchor=base west,
                parent anchor=east,
                child anchor=west,
                node options={align=center},
                align = center,
                base=left,
                font=\small,
                rectangle,
                draw=hidden-draw,
                rounded corners,
                edge+={darkgray, line width=1pt},
                s sep=3pt,
                inner xsep=2pt,
                inner ysep=3pt,
                ver/.style={rotate=90, child anchor=north, parent anchor=south, anchor=center},
            },
            where level=1{text width=5.0em,font=\scriptsize}{},
            where level=2{text width=5.6em,font=\scriptsize}{},
            where level=3{text width=6.8em,font=\scriptsize}{},
            [
            DRL Explanation Methods, ver
            [
            Feature-level
            [
            Perturbation-based
                [
               ~\cite{zahavy2016graying,greydanus2018visualizing, atrey2019exploratory}
                , leaf, text width=20em
                ]
            ]
            [
            Gradient-based
                [
               ~\cite{wang2016dueling,selvaraju2017grad,sundararajan2017axiomatic}
                , leaf, text width=21em
                ]
            ]
            [
            Attention-based
                [
               ~\cite{mott2019towards,nikulin2019free}
                , leaf, text width=12em
                ]
            ]
            ]
            [
            State-level
                [
            Offline \\ Trajectories
                [
               ~\cite{guo2021edge,yu2023airs,liu2023learning}
                , leaf, text width=16em
                ]
                ]
                [
            Online \\ Interactions
                [
               ~\cite{jacq2022lazymdp,cheng2023statemask,chengrice}
                , leaf, text width=14em
                ]
                ]
            ]
            [
            Dataset-level
                [
            Influence \\ Functions
                [
               ~\cite{koh2017understanding,li2024influence,matelsky2024empirical,ruis2024procedural}
                , leaf, text width=24em
                ]
                ]
                [
            Data Shapley
                [
               ~\cite{ghorbani2019data,wanghelpful, schoch2023srw}
                , leaf, text width=20em
                ]
                ]
                [
            Data Masking
                [
               ~\cite{dong2024promptexp, lin2024data}
                , leaf, text width=11em
                ]
                ]
            ]
            [
            Model-level
                [
            Transparent \\ Architectures
                [
               ~\cite{topin2021iterative,ding2020cdt, demircan2024sparse}
                , leaf, text width=19em
                ]
                ]
                [
            Rule \\ Extraction
                [
               ~\cite{soares2020explaining,likmeta2020combining}
                , leaf, text width=13em
                ]
                ]
            ]
            ]
        \end{forest}
    }
\caption{Taxonomy of DRL Explanation Methods}
\label{fig:drl_explanation}
\end{figure*}
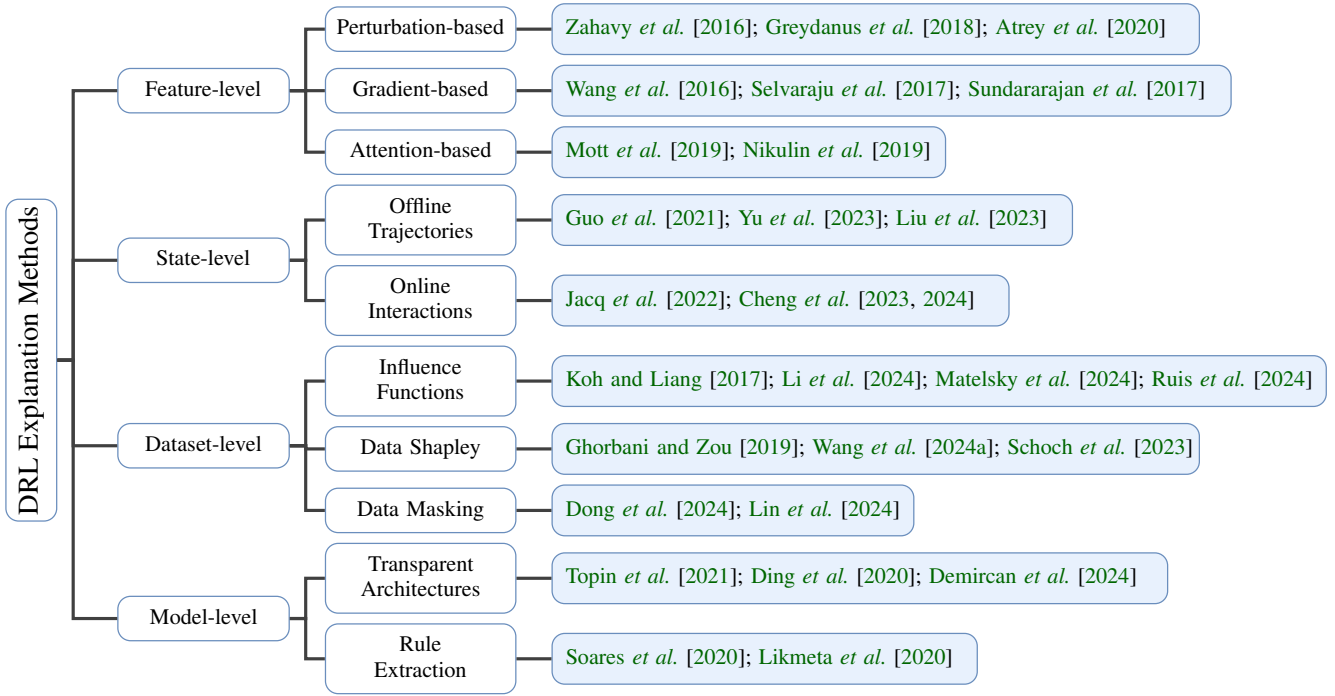

Existing approaches to explaining deep reinforcement learning can be broadly categorized into four categories: \circled{1} \textbf{Feature-level Explanation Methods}, which focuses on pinpointing the most important feature in the DRL agent's observation; \circled{2} \textbf{State-level Explanation Methods}, which identifies the most critical steps in the RL trajectory; \circled{3} \textbf{Dataset-level Explanation Methods}, which selects the most influential data in RL; \circled{4} \textbf{Model-level Explanation Methods}, which focuses on the self-explainability of RL policy models. A summary of selected methods is provided in \autoref{fig:drl_explanation}.

\subsection{Feature-level Explanation Methods}
Feature-level explanation methods aim to identify the most important features in the agent's observation space that influence its decision-making. These methods are particularly useful for understanding how an agent processes visual inputs. 

\citet{zahavy2016graying} approximated the behavior of DRL agents via Semi-Aggregated Markov Decision Processes (SAMDPs) and analyzed the high-level temporal structure of the policy with the more interpretable SAMDPs. However, the explanation from SAMDPs is drawn from t-SNE clusters which could be uninformative for users without machine learning backgrounds. To make the explanation more accessible, \citet{greydanus2018visualizing} proposed a feature-level explanation method to visualize the importance of pixels in Atari game frames by perturbing the input and observing changes in the agent's policy. 

In addition to perturbation-based saliency methods, some researchers also proposed gradient-based saliency methods that use gradients of the agent's policy or value function to pinpoint the most important feature in DRL agent's observation. \citet{wang2016dueling} extend gradient-based saliency maps to deep RL by computing the Jacobian of the output logits with respect to a stack of input images. \citet{joo2019visualization} leveraged Grad-Cam~\citep{selvaraju2017grad} to visualize the important features towards the DRL agent's behavior. \citet{chengrice} mentioned that we can also use integrated gradients~\citep{sundararajan2017axiomatic} to identify the most important features.

Recent advancements in deep reinforcement learning (DRL) also introduce attention-based mechanisms to enhance feature-level explanations of agent behavior. These methods aim to improve interpretability by enabling agents to focus on task-relevant information within their observation space. For instance, \citet{mott2019towards} proposed an attention-augmented agent that employs a soft attention mechanism, allowing the agent to sequentially query its environment and focus on pertinent features during decision-making. This approach not only enhances performance but also provides interpretable attention maps that highlight the areas of the input contributing to the agent's actions. Similarly, \citet{nikulin2019free} introduced a method that integrates an attention module into the agent's architecture, producing saliency maps that visualize the importance of different input regions in the agent's decision process. 

These methods provide insights into the agent's perception of the environment but are often limited to explaining low-level features rather than high-level decision-making processes.

\subsection{State-level Explanation Methods}
State-level explanation methods focus on identifying critical states in the agent's trajectory that significantly impact its performance. These methods are useful for understanding the agent's behavior over time and diagnosing failures. We categorize state-level explanation methods into two categories: (1) Explain through offline trajectories; (2) Explain through online interactions.

For the first category, \cite{guo2021edge} first proposed EDGE that establishes state-reward relationship
by collecting a set of trajectories and then approximating an explanation model offline with the Gaussian Process. Note that, EDGE provides a {\em global explanation} for the policy network. AIRS~\citep{yu2023airs} further introduces a {\em local explanation} method to identify critical time steps for a given trajectory of interest. AIRS pre-collects a set of trajectories and utilizes a deep neural network to estimate the contribution of each state to the final rewards for each trajectory. \cite{liu2023learning} proposed a Deep State Identifier that learns to predict returns from episodes and uses mask-based sensitivity analysis to extract important states. However, the fidelity of these methods is highly related to the quality of the pre-collected trajectories, which limits their ability to measure the importance of ``unseen states''. 

For the second category, \citet{jacq2022lazymdp} presented LazyMDP, which extends the action space with a lazy action and learns to switch between the default action and the lazy action. The states where the policy diverges from the default are further interpreted as non-important states. \cite{cheng2023statemask} proposed StateMask, which online trains a mask network in parallel with the agent's policy network. The mask network learns to ``blind'' the agent's observations at certain time steps (by taking random actions) while minimizing the impact of blinding to the final reward. The time steps when the agent could be blinded are identified as non-critical steps. 

State-level explanations are particularly valuable for debugging and improving RL agents, as they highlight the most influential moments in the agent's decision-making process.

\subsection{Dataset-level Explanation Methods}
Dataset-level explanation methods focus on understanding how specific training examples influence the learned policy of an RL agent. By identifying which data points have the most impact on the policy updates, researchers and practitioners can better diagnose training inefficiencies, detect harmful experiences, and refine data collection strategies. Recent work has highlighted multiple approaches for quantifying this influence:

\paragraph{Influence Functions.} Originally introduced by \citet{koh2017understanding}, influence functions estimate how an upweighting or removal of a single training example impacts model parameters. In RL contexts, these techniques can be adapted to analyze individual experiences in a replay buffer, thereby revealing which transitions most critically shape the agent’s behavior. When incorporating RL with LLMs, \citet{li2024influence,matelsky2024empirical,ruis2024procedural} also investigated the feasibility of leveraging influence functions to identify influential data. However, they found that influence functions show poor performance and the reasons might be (1) inevitable approximation errors when estimating the inverse-Hessian vector products (iHVP) component due to the scale of LLMs, (2) uncertain convergence during fine-tuning, (3) the definition of influential data as changes in model parameters do not necessarily correlate with changes in LLM behavior.

\paragraph{Data Shapley Values.} Shapley values, proposed by \citet{ghorbani2019data}, offer a game-theoretic metric for attributing credit to each data point. By considering all possible subsets of the training set, Data Shapley Values can rank experiences according to their overall contribution to policy performance. However, the original Data Shapley Values are computationally intensive, \citet{wanghelpful} proposed an approximation method FreeShap for instance attribution based on the neural tangent kernel, which makes this method feasible for explaining LLM predictions.

\paragraph{Data Masking.} Recent advances have introduced masking as a way to figure out how specific elements of a training dataset shape an agent’s learning process \citep{dong2024promptexp, lin2024data}. Rather than simply omitting entire experiences, data masking strategically hides or perturbs certain tokens and observes how these modifications affect the LLM's performance. Therefore, researchers can pinpoint the data components most critical to LLM training and can construct a pruned dataset based on the critical data to efficiently train a LLM. 

Dataset-level explanations help researchers and practitioners understand the role of training data in shaping the RL agent's behavior and can guide the design of more efficient and effective training schemes.

\subsection{Model-level Explanation Methods}
Model-level explanation methods focus on the self-explainability of RL policy models, aiming to make the agent's decision-making process inherently interpretable. These methods often involve designing transparent architectures (\eg decision tree~\citep{topin2021iterative, ding2020cdt}) or extracting human-understandable rules~\citep{soares2020explaining, likmeta2020combining} from the agent's policy. 
\cite{demircan2024sparse} utilize sparse auto-encoders within the policy network to provide detailed explanations of LLM's behavior, specifically focusing on how the network approximates Q-learning by revealing the underlying structure and decision-making process of the model.

Model-level explanations are particularly valuable for applications requiring high transparency, such as healthcare and autonomous driving, where understanding the agent's reasoning is critical for trust and safety.
\section{Measuring XRL}
Evaluating the quality of explanations in RL requires a multi-faceted approach that captures both user-centered dimensions and objective metrics. This section outlines two broad categories of assessment, \ie qualitative and quantitative.

\subsection{Qualitative Evaluation}
\textbf{Interpretability and Clarity.}
At the heart of XRL is the need for explanations that humans find meaningful and intuitive. Qualitative evaluation often begins with user studies, such as surveys, to gauge how well participants understand the explanation and whether the information provided is perceived as coherent and sufficient for understanding policy decisions. Most researchers provide a visualization of the proposed explanation technique to demonstrate to the participants that the explanation can help them understand the DRL agent's behavior. For feature-level explanations, \citet{greydanus2018visualizing} generated saliency videos to show the feature-level explanations for Atari games and conducted a survey over 31 students at Oregon State University to measure how their visualization helps non-experts with these Atari games. For state-level explanations, \citet{cheng2023statemask} generated game trajectories with a color bar behind each frame to indicate the importance of each state and invited participants to answer a questionnaire to demonstrate their method StateMask could help humans gain a better understanding
of a DRL agent’s behavior.

\noindent\textbf{User-Centered Design Considerations.}
The qualitative evaluation also informs iterative refinement of explanation interfaces. By examining user reactions, researchers and designers can identify which presentation formats (e.g., visual overlays, textual rationales, or example-based justifications) are most effective. This feedback loop, encompassing pilot testing and usability reviews, ensures that explanations remain aligned with the domain’s practical needs and the target audience’s expertise.

\subsection{Quantitative Evaluation}
\textbf{Fidelity and Faithfulness.}
A key quantitative metric is how closely an explanation reflects the true policy or behavior of the RL agent. To evaluate the fidelity of the explanation in RL, researchers commonly use a perturbation-based approach~\citep{guo2021edge,cheng2023statemask}. The researchers remove features/states/data points identified as critical in the explanation and check if such a removal substantially degrades the agent’s performance. A dual form of this approach is to remove the non-critical features/states/data points and the agent's performance is expected to have limited difference. The fidelity score is further measured as the performance difference of the DRL agent before and after perturbing a fixed number of pixels/states/data points. When perturbing the same number of (critical) pixels/states/data points, a higher performance difference indicates a higher fidelity of the explanation method.

\textbf{Downstream Performance Impact.}
XRL systems can also be evaluated on whether their explanations enhance agent performance. For instance, \citet{chengrice} tested their proposed refining method based on the critical steps identified by different explanation methods and compared the agent’s performance after refining to evaluate the quality of these explanation methods.
\section{Applications of Explanations}
With the explanation of RL, there can be different applications of it - they can be leveraged both constructively (for policy refinement and debugging) and potentially destructively (for launching adversarial attacks). These applications demonstrate how fidelity and interpretability impact the effectiveness of explanation-based interventions in real-world scenarios.

\begin{table*}[ht]
\centering
\resizebox{2.0\columnwidth}{!}{
\begin{tabular}{lll}
\toprule
\textbf{Category} & \textbf{Subcategory} & \textbf{Citation} \\
\midrule
\multirow{1}{*}{Launching Adversarial Attacks} 
 & Targeted Attack & \citep{lin2020robustness,guo2021edge, cheng2023statemask, wang2024rlhfpoison} \\
\midrule
\multirow{2}{*}{Mitigating Adversarial Attacks} 
 & Blinding Observations & \citep{guo2021edge} \\
 & Shielding Backdoor Triggers & \citep{yuan2024shine} \\
\midrule
\multirow{2}{*}{Policy Refinement} 
 & Human-in-the-Loop Correction & \citep{van2022correct, jiang2024reinforcement} \\
 & Automated Policy Refinement & \citep{guo2021edge, cheng2023statemask, yu2023airs, chengrice,liu2025utilizing} \\
\bottomrule
\end{tabular}
}
\caption{Taxonomy of Explanation-based Interventions in DRL.}
\label{tab:taxonomy}
\end{table*}

\subsection{Launching Adversarial Attacks}
Recent work demonstrates that explanations of a DRL agent's policy can be repurposed to compromise the agent's performance. Recent studies have revealed that explanations of a Deep Reinforcement Learning (DRL) agent's policy can be exploited to compromise the agent's performance. For instance, \citet{lin2020robustness} demonstrated the vulnerability of cooperative Multi-Agent Reinforcement Learning systems to adversarial attacks by introducing perturbations based on feature-level explanations (\ie saliency) to the state space. They proposed a mechanism where an adversary adds perturbations to the observations of a single agent within a team, leading to a significant decrease in overall team performance.

Besides leveraging feature-level explanations to launch adversarial attacks, researchers also demonstrate that state-level explanations can be utilized to attack DRL agents. EDGE \citep{guo2021edge} proposes a more targeted approach by leveraging explanations to identify critical time steps during an episode. The attacker first collects winning episodes from the victim agent and uses post-hoc explanations to highlight moments where actions strongly contribute to victory. By forcing the agent to take sub-optimal actions at these identified crucial steps, the attack achieves significant performance degradation with minimal intervention.

Subsequent research by \citet{cheng2023statemask} confirms this explanation-driven attack generalizes across different DRL environments, showing that targeting just 10\% of time steps can substantially reduce agent reward. Notably, attacks guided by high-fidelity explanation methods prove more effective than those using lower-fidelity alternatives, highlighting how better interpretability tools can paradoxically increase vulnerability.

In addition to exploiting feature-level and state-level explanations, recent research has explored the use of dataset-level explanations to launch adversarial attacks on LLMs. A study by \citet{wang2024rlhfpoison} investigates the vulnerabilities of reinforcement learning with human feedback. The researchers employ a gradient-based dataset-level explanation method to identify influential data points within the training set. By poisoning a small percentage of critical data, an adversary can significantly manipulate the LLM's behavior, leading to the elicitation of harmful responses. 

This line of work highlights the dual-edged nature of interpretability in RL. While explanations are invaluable for debugging and understanding agent behavior, they can also expose vulnerabilities. By identifying specific moments when an agent's correct actions matter most, adversaries can focus on minimal but high-impact interventions. Consequently, researchers must carefully consider the security implications of providing public or easily accessible explanation systems, especially in safety-critical or competitive domains.

\subsection{Mitigating Adversarial Attacks}

XRL methods not only reveal how adversaries can manipulate agents but can also guide the design of robust policies. By pinpointing which states or actions are most vulnerable, developers can selectively limit or modify the agent's observations and decision pathways at crucial moments, ultimately reducing susceptibility to adversarial inputs.

\textbf{Blinding Observations at Critical Time Steps.}
\citet{guo2021edge} illustrated how explanations of the victim agent's losing episodes in the You-Shall-Not-Pass game~\citep{todorov2012mujoco} uncover the specific times when adversarial actions (\eg, pretending to fall) most effectively mislead the agent. By analyzing contrastive explanations—comparing losing and winning trajectories—it becomes clear that the agent's focus on adversarial cues at certain time steps can trigger suboptimal responses. The authors proposed ``blinding'' the victim agent to these cues precisely at those critical moments. Experimental results show that this explanation-driven defense significantly boosts the victim's win rate, highlighting how identifying the root cause of agent failures can lead to targeted and effective countermeasures.

\textbf{Detecting and Shielding Backdoor Triggers.}
Another line of work focuses on a subtler attack vector: maliciously injected backdoors. \citet{yuan2024shine} introduced \textit{SHINE}, a method to shield a pre-trained agent from both perturbation-based and adversarial-agent attacks in a poisoned environment. SHINE first gathers trajectories and employs a two-stage explanation process to (1) locate states where a backdoor trigger is likely active and (2) isolate the common subset of features critical to the agent's decisions in those states. These features are then treated as the backdoor signature. In the second stage, SHINE retrains the policy to neutralize the trigger's influence while preserving performance in a clean environment. This careful mixture of explanation and policy adjustment provides theoretical guarantees of improved robustness.

These defense mechanisms highlight how explanations serve defensive purposes in adversarial contexts. By precisely identifying \emph{where} and \emph{how} an agent's decision-making is compromised, explanation-guided strategies enable targeted fixes that enhance robustness. This demonstrates that transparency, when properly leveraged, can be a powerful tool for securing DRL agents rather than just exposing their vulnerabilities.

\subsection{Policy Refinement Through Explanations}

To refine the policy of the agents, conventional methods such as continual training~\citep{fickinger2021scalable} often fall short due to a lack of knowledge of the root causes of errors. There are two categories of methods for policy refinement through explanations:
\begin{itemize}
    \item \textbf{Human-in-the-Loop Correction:} Domain experts or non-experts identify suboptimal actions or critical states, providing corrective demonstrations or reward adjustments.
    \item \textbf{Automated Policy Refinement with Explanation:} Explanation techniques automatically identify pivotal states and refine the target agent's policy based on the explanation.
\end{itemize}

For the first category, \cite{van2022correct} proposed to utilize human feedback to correct the agent's failures. 
More specifically, when the agent fails, humans (can be non-experts) are involved to point out how to avoid such a failure (\ie what action should be done instead, and what action should be forbidden). Based on human feedback, the DRL agent gets retrained by taking the human-refined action in those important time steps and finally obtains the corrected policy. 
The downside is that it relies on humans to identify critical steps and craft rules for alternative actions. This can be challenging for a large action space, and the retraining process is ad-hoc and time-consuming. To address the challenges of imperfect corrective actions and extensive human labor, \citet{jiang2024reinforcement} introduced the Iterative learning from Corrective actions and Proxy rewards (ICoPro) framework. In this approach, human labelers provide corrective actions on the agent's trajectories, which are then incorporated into the Q-function using a margin loss to enforce adherence to the labeler's preferences. The agent undergoes iterative training, balancing learning from both proxy rewards and human feedback. Notably, ICoPro integrates pseudo-labels from the target Q-network to reduce human labor and stabilize training. Experimental results in various tasks, including Atari games and autonomous driving scenarios, demonstrate that ICoPro effectively aligns agent behavior with human preferences, even when both proxy rewards and corrective actions are imperfect.

For the second category, \cite{guo2021edge} proposed an explanation-guided policy refinement approach to automatically correct policy errors without relying on explicit human feedback. Their method first identifies losing episodes of the target agent and pinpoints crucial time steps within those episodes using its proposed explanation technique. The authors employ a fixed number of random explorations at the identified critical time steps. Any random actions that transform a losing episode into a win get stored in a look-up table as a remediation policy.
When deployed, the agent consults this table at run-time: if the current state matches one of the stored entries, the agent applies the corresponding remediation action; otherwise, it defaults to its original policy. The success of this policy refinement approach depends heavily on the budget of random exploration and the size of the look-up table. \cite{cheng2023statemask, yu2023airs} further proposed to use DRL explanation methods to identify critical time steps and refine the agent by resetting the environment to the critical states and subsequently resuming training the DRL agents from these critical states. However, this refining strategy can easily lead to overfitting as evidenced in \cite{chengrice} and cannot help the agent escape the local optimal. 
\cite{chengrice} further proposed a novel refining strategy to construct a mixed initial state distribution with both the identified critical states and the default initial states to avoid overfitting and encourage the agent to perform exploration during the refining process. Recently, \citet{liu2025utilizing} proposed a novel framework that leverages explainable reinforcement learning (XRL) to enhance policy refinement. This approach addresses the challenges of DRL agents' lack of transparency and suboptimal performance by providing a two-level explanation of the agents' decision-making processes. The framework identifies mistakes made by the DRL agent and formulates a constrained bi-level optimization problem to learn how to best utilize these explanations for policy improvement. The upper level of the optimization learns how to use high-level explanations to shape the reward function, while the lower level solves a constrained RL problem using low-level explanations. The proposed algorithm theoretically guarantees global optimality and has demonstrated superior performance in MuJoCo experiments compared to state-of-the-art baselines.
\section{Conclusion and Future Directions}

This survey has reviewed recent advances in the field of XRL, emphasizing a range of techniques - from feature-level, and state-level to dataset-level approaches, and illustrating their roles in adversarial attacks and mitigation, and policy refinement. Evidence across these methods indicates that effective explanations can significantly enhance trust and debugging efficiency in real-world deployments of deep reinforcement learning. Nonetheless, substantial gaps remain to be addressed, which are summarized as follows.

\paragraph{User-Oriented Explanations.}
Although existing techniques could highlight critical features/states to illustrate an agent’s decision-making process, these granular depictions can be difficult for non-expert users to interpret. In the case of critical features, users who lack domain knowledge (\eg specific familiarity with a particular game environment) may struggle to grasp the significance of highlighted features and how they influence the agent’s actions. Meanwhile, understanding critical states often demands that users examine multiple visual frames and then manually summarize what these states imply about the agent’s strategy. This process can be cognitively taxing, as it requires piecing together dispersed information and inferring the agent’s underlying rationale without clear contextual guidance.

To address these challenges, future research should therefore prioritize strategy-level or narrative-based explanations, which can provide higher-level rationales that are more accessible to general audiences. In particular, leveraging vision–language models or other multimodal architectures could facilitate the presentation of natural language narratives that encapsulate an agent’s overarching goals, strategies, and reasoning. These narrative formats have the potential to reduce cognitive load, enabling end users to more intuitively comprehend and trust the agent’s behavior.

\paragraph{Developer-Oriented Explanations. }
In contrast, developers and researchers frequently require detailed insights into an agent’s decision-making process. Mechanistic interpretation methods, such as sparse autoencoders or network dissection, could illuminate hidden representations and policy structures. These more granular approaches enable targeted policy debugging by pinpointing design flaws or overfitting at the architectural level. Crucially, explanations for developers should be \emph{actionable}, which could be compatible with policy refinement workflows to accelerate iterative improvements.

In addition to improving interpretability, explainability tools offer considerable potential for enhancing policy performance. For instance, in game-theoretic contexts, explanations can help identify equilibrium strategies or support robust multi-agent interactions. In hierarchical reinforcement learning, clarifying subtask transitions can streamline learning in sparse-reward or long-horizon tasks. Similarly, in curriculum learning, highlighting critical states through explanation techniques can aid developers in selecting more effective initial conditions. Moving forward, future research should focus on aligning these interpretability and performance objectives by examining how transparent representations of policy decisions can foster robust learning or facilitate agent learning.

\bibliographystyle{named}
\bibliography{ijcai25}

\begin{thebibliography}{}

\bibitem[\protect\citeauthoryear{Atrey \bgroup \em et al.\egroup
  }{2020}]{atrey2019exploratory}
Akanksha Atrey, Kaleigh Clary, and David Jensen.
\newblock Exploratory not explanatory: Counterfactual analysis of saliency maps
  for deep reinforcement learning.
\newblock In {\em Proc. of ICLR}, 2020.

\bibitem[\protect\citeauthoryear{Bai \bgroup \em et al.\egroup
  }{2022}]{bai2022training}
Yuntao Bai, Andy Jones, Kamal Ndousse, Amanda Askell, Anna Chen, Nova DasSarma,
  Dawn Drain, Stanislav Fort, Deep Ganguli, Tom Henighan, et~al.
\newblock Training a helpful and harmless assistant with reinforcement learning
  from human feedback.
\newblock {\em arXiv preprint arXiv:2204.05862}, 2022.

\bibitem[\protect\citeauthoryear{Cheng \bgroup \em et al.\egroup
  }{2023}]{cheng2023statemask}
Zelei Cheng, Xian Wu, Jiahao Yu, Wenhai Sun, Wenbo Guo, and Xinyu Xing.
\newblock Statemask: Explaining deep reinforcement learning through state mask.
\newblock In {\em Proc. of NeurIPS}, 2023.

\bibitem[\protect\citeauthoryear{Cheng \bgroup \em et al.\egroup
  }{2024}]{chengrice}
Zelei Cheng, Xian Wu, Jiahao Yu, Sabrina Yang, Gang Wang, and Xinyu Xing.
\newblock Rice: Breaking through the training bottlenecks of reinforcement
  learning with explanation.
\newblock In {\em Proc. of ICML}, 2024.

\bibitem[\protect\citeauthoryear{Demircan \bgroup \em et al.\egroup
  }{2024}]{demircan2024sparse}
Can Demircan, Tankred Saanum, Akshay~K Jagadish, Marcel Binz, and Eric Schulz.
\newblock Sparse autoencoders reveal temporal difference learning in large
  language models.
\newblock {\em arXiv preprint arXiv:2410.01280}, 2024.

\bibitem[\protect\citeauthoryear{Ding \bgroup \em et al.\egroup
  }{2020}]{ding2020cdt}
Zihan Ding, Pablo Hernandez-Leal, Gavin~Weiguang Ding, Changjian Li, and
  Ruitong Huang.
\newblock Cdt: Cascading decision trees for explainable reinforcement learning.
\newblock {\em arXiv preprint arXiv:2011.07553}, 2020.

\bibitem[\protect\citeauthoryear{Dong \bgroup \em et al.\egroup
  }{2023}]{dong2023raft}
Hanze Dong, Wei Xiong, Deepanshu Goyal, Yihan Zhang, Winnie Chow, Rui Pan,
  Shizhe Diao, Jipeng Zhang, KaShun SHUM, and Tong Zhang.
\newblock {RAFT}: Reward ranked finetuning for generative foundation model
  alignment.
\newblock {\em Transactions on Machine Learning Research}, 2023.

\bibitem[\protect\citeauthoryear{Dong \bgroup \em et al.\egroup
  }{2024}]{dong2024promptexp}
Ximing Dong, Shaowei Wang, Dayi Lin, Gopi~Krishnan Rajbahadur, Boquan Zhou,
  Shichao Liu, and Ahmed~E Hassan.
\newblock Promptexp: Multi-granularity prompt explanation of large language
  models.
\newblock {\em arXiv preprint arXiv:2410.13073}, 2024.

\bibitem[\protect\citeauthoryear{Feriani and Hossain}{2021}]{feriani2021single}
Amal Feriani and Ekram Hossain.
\newblock Single and multi-agent deep reinforcement learning for ai-enabled
  wireless networks: A tutorial.
\newblock {\em IEEE Communications Surveys \& Tutorials}, 23(2):1226--1252,
  2021.

\bibitem[\protect\citeauthoryear{Fickinger \bgroup \em et al.\egroup
  }{2021}]{fickinger2021scalable}
Arnaud Fickinger, Hengyuan Hu, Brandon Amos, Stuart Russell, and Noam Brown.
\newblock Scalable online planning via reinforcement learning fine-tuning.
\newblock In {\em Proc. of NeurIPS}, 2021.

\bibitem[\protect\citeauthoryear{Ghorbani and Zou}{2019}]{ghorbani2019data}
Amirata Ghorbani and James Zou.
\newblock Data shapley: Equitable valuation of data for machine learning.
\newblock In {\em Proc. of ICML}, 2019.

\bibitem[\protect\citeauthoryear{Greydanus \bgroup \em et al.\egroup
  }{2018}]{greydanus2018visualizing}
Samuel Greydanus, Anurag Koul, Jonathan Dodge, and Alan Fern.
\newblock Visualizing and understanding atari agents.
\newblock In {\em Proc. of ICML}, 2018.

\bibitem[\protect\citeauthoryear{Guo \bgroup \em et al.\egroup
  }{2021}]{guo2021edge}
Wenbo Guo, Xian Wu, Usmann Khan, and Xinyu Xing.
\newblock Edge: Explaining deep reinforcement learning policies, 2021.

\bibitem[\protect\citeauthoryear{Jacq \bgroup \em et al.\egroup
  }{2022}]{jacq2022lazymdp}
Alexis Jacq, Johan Ferret, Olivier Pietquin, and Matthieu Geist.
\newblock Lazy-mdps: Towards interpretable rl by learning when to act.
\newblock In {\em Proc. of AAMAS}, 2022.

\bibitem[\protect\citeauthoryear{Jiang \bgroup \em et al.\egroup
  }{2024}]{jiang2024reinforcement}
Zhaohui Jiang, Xuening Feng, Paul Weng, Yifei Zhu, Yan Song, Tianze Zhou,
  Yujing Hu, Tangjie Lv, and Changjie Fan.
\newblock Reinforcement learning from imperfect corrective actions and proxy
  rewards.
\newblock {\em arXiv preprint arXiv:2410.05782}, 2024.

\bibitem[\protect\citeauthoryear{Joo and Kim}{2019}]{joo2019visualization}
Ho-Taek Joo and Kyung-Joong Kim.
\newblock Visualization of deep reinforcement learning using grad-cam: how ai
  plays atari games?
\newblock In {\em Proc. of CoG}, 2019.

\bibitem[\protect\citeauthoryear{Kaiser \bgroup \em et al.\egroup
  }{2020}]{kaiser2020model}
{\L}ukasz Kaiser, Mohammad Babaeizadeh, Piotr Mi{\l}os, B{\l}a{\.z}ej
  Osi{\'n}ski, Roy~H Campbell, Konrad Czechowski, Dumitru Erhan, Chelsea Finn,
  Piotr Kozakowski, Sergey Levine, et~al.
\newblock Model-based reinforcement learning for atari.
\newblock In {\em Proc. of ICLR}, 2020.

\bibitem[\protect\citeauthoryear{Kalashnikov \bgroup \em et al.\egroup
  }{2018}]{kalashnikov2018scalable}
Dmitry Kalashnikov, Alex Irpan, Peter Pastor, Julian Ibarz, Alexander Herzog,
  Eric Jang, Deirdre Quillen, Ethan Holly, Mrinal Kalakrishnan, Vincent
  Vanhoucke, et~al.
\newblock Scalable deep reinforcement learning for vision-based robotic
  manipulation.
\newblock In {\em Proc. of CoRL}, 2018.

\bibitem[\protect\citeauthoryear{Koh and Liang}{2017}]{koh2017understanding}
Pang~Wei Koh and Percy Liang.
\newblock Understanding black-box predictions via influence functions.
\newblock In {\em Proc. of ICML}, 2017.

\bibitem[\protect\citeauthoryear{Li \bgroup \em et al.\egroup
  }{2024}]{li2024influence}
Zhe Li, Wei Zhao, Yige Li, and Jun Sun.
\newblock Do influence functions work on large language models?
\newblock {\em arXiv preprint arXiv:2409.19998}, 2024.

\bibitem[\protect\citeauthoryear{Likmeta \bgroup \em et al.\egroup
  }{2020}]{likmeta2020combining}
Amarildo Likmeta, Alberto~Maria Metelli, Andrea Tirinzoni, Riccardo Giol,
  Marcello Restelli, and Danilo Romano.
\newblock Combining reinforcement learning with rule-based controllers for
  transparent and general decision-making in autonomous driving.
\newblock {\em Robotics and Autonomous Systems}, 131:103568, 2020.

\bibitem[\protect\citeauthoryear{Lin \bgroup \em et al.\egroup
  }{2020}]{lin2020robustness}
Jieyu Lin, Kristina Dzeparoska, Sai~Qian Zhang, Alberto Leon-Garcia, and
  Nicolas Papernot.
\newblock On the robustness of cooperative multi-agent reinforcement learning.
\newblock In {\em Proc. of IEEE S\&P Workshop}, 2020.

\bibitem[\protect\citeauthoryear{Lin \bgroup \em et al.\egroup
  }{2024}]{lin2024data}
Xinyu Lin, Wenjie Wang, Yongqi Li, Shuo Yang, Fuli Feng, Yinwei Wei, and
  Tat-Seng Chua.
\newblock Data-efficient fine-tuning for llm-based recommendation.
\newblock In {\em Proc. of SIGIR}, 2024.

\bibitem[\protect\citeauthoryear{Liu and Zhu}{2025}]{liu2025utilizing}
Shicheng Liu and Minghui Zhu.
\newblock Utilizing explainable reinforcement learning to improve reinforcement
  learning: A theoretical and systematic framework.
\newblock In {\em Proc. of ICLR}, 2025.

\bibitem[\protect\citeauthoryear{Liu \bgroup \em et al.\egroup
  }{2023}]{liu2023learning}
Haozhe Liu, Mingchen Zhuge, Bing Li, Yuhui Wang, Francesco Faccio, Bernard
  Ghanem, and J{\"u}rgen Schmidhuber.
\newblock Learning to identify critical states for reinforcement learning from
  videos.
\newblock In {\em Proc. of ICCV}, 2023.

\bibitem[\protect\citeauthoryear{Liu \bgroup \em et al.\egroup
  }{2024}]{liu2024dynamic}
Xiao-Yang Liu, Ziyi Xia, Hongyang Yang, Jiechao Gao, Daochen Zha, Ming Zhu,
  Christina~Dan Wang, Zhaoran Wang, and Jian Guo.
\newblock Dynamic datasets and market environments for financial reinforcement
  learning.
\newblock {\em Machine Learning}, 113(5):2795--2839, 2024.

\bibitem[\protect\citeauthoryear{Matelsky \bgroup \em et al.\egroup
  }{2024}]{matelsky2024empirical}
Jordan~K Matelsky, Lyle Ungar, and Konrad~P Kording.
\newblock Empirical influence functions to understand the logic of fine-tuning.
\newblock {\em arXiv preprint arXiv:2406.00509}, 2024.

\bibitem[\protect\citeauthoryear{Mnih \bgroup \em et al.\egroup
  }{2015}]{mnih2015human}
Volodymyr Mnih, Koray Kavukcuoglu, David Silver, Andrei~A Rusu, Joel Veness,
  Marc~G Bellemare, Alex Graves, Martin Riedmiller, Andreas~K Fidjeland, Georg
  Ostrovski, et~al.
\newblock Human-level control through deep reinforcement learning.
\newblock {\em Nature}, 518(7540):529--533, 2015.

\bibitem[\protect\citeauthoryear{Mnih}{2013}]{mnih2013playing}
Volodymyr Mnih.
\newblock Playing atari with deep reinforcement learning.
\newblock {\em arXiv preprint arXiv:1312.5602}, 2013.

\bibitem[\protect\citeauthoryear{Mott \bgroup \em et al.\egroup
  }{2019}]{mott2019towards}
Alexander Mott, Daniel Zoran, Mike Chrzanowski, Daan Wierstra, and Danilo
  Jimenez~Rezende.
\newblock Towards interpretable reinforcement learning using attention
  augmented agents.
\newblock In {\em Proc. of NeurIPS}, 2019.

\bibitem[\protect\citeauthoryear{Nikulin \bgroup \em et al.\egroup
  }{2019}]{nikulin2019free}
Dmitry Nikulin, Anastasia Ianina, Vladimir Aliev, and Sergey Nikolenko.
\newblock Free-lunch saliency via attention in atari agents.
\newblock In {\em Proc. of ICCV Workshop}, 2019.

\bibitem[\protect\citeauthoryear{Ouyang \bgroup \em et al.\egroup
  }{2022}]{ouyang2022training}
Long Ouyang, Jeffrey Wu, Xu~Jiang, Diogo Almeida, Carroll Wainwright, Pamela
  Mishkin, Chong Zhang, Sandhini Agarwal, Katarina Slama, Alex Ray, et~al.
\newblock Training language models to follow instructions with human feedback.
\newblock In {\em Proc. of NeurIPS}, 2022.

\bibitem[\protect\citeauthoryear{Rafailov \bgroup \em et al.\egroup
  }{2023}]{rafailov2024direct}
Rafael Rafailov, Archit Sharma, Eric Mitchell, Christopher~D Manning, Stefano
  Ermon, and Chelsea Finn.
\newblock Direct preference optimization: Your language model is secretly a
  reward model.
\newblock In {\em Proc. of NeurIPS}, 2023.

\bibitem[\protect\citeauthoryear{Ruis \bgroup \em et al.\egroup
  }{2024}]{ruis2024procedural}
Laura Ruis, Maximilian Mozes, Juhan Bae, Siddhartha~Rao Kamalakara, Dwarak
  Talupuru, Acyr Locatelli, Robert Kirk, Tim Rockt{\"a}schel, Edward
  Grefenstette, and Max Bartolo.
\newblock Procedural knowledge in pretraining drives reasoning in large
  language models.
\newblock {\em arXiv preprint arXiv:2411.12580}, 2024.

\bibitem[\protect\citeauthoryear{Schoch \bgroup \em et al.\egroup
  }{2023}]{schoch2023srw}
Stephanie Schoch, Ritwick Mishra, and Yangfeng Ji.
\newblock Data selection for fine-tuning large language models using
  transferred shapley values.
\newblock In {\em Proc. of ACL}, 2023.

\bibitem[\protect\citeauthoryear{Schulman \bgroup \em et al.\egroup
  }{2017}]{schulman2017proximal}
John Schulman, Filip Wolski, Prafulla Dhariwal, Alec Radford, and Oleg Klimov.
\newblock Proximal policy optimization algorithms.
\newblock {\em arXiv preprint arXiv:1707.06347}, 2017.

\bibitem[\protect\citeauthoryear{Selvaraju \bgroup \em et al.\egroup
  }{2017}]{selvaraju2017grad}
Ramprasaath~R Selvaraju, Michael Cogswell, Abhishek Das, Ramakrishna Vedantam,
  Devi Parikh, and Dhruv Batra.
\newblock Grad-cam: Visual explanations from deep networks via gradient-based
  localization.
\newblock In {\em Proc. of ICCV}, 2017.

\bibitem[\protect\citeauthoryear{Silver \bgroup \em et al.\egroup
  }{2017}]{silver2017mastering}
David Silver, Julian Schrittwieser, Karen Simonyan, Ioannis Antonoglou, Aja
  Huang, Arthur Guez, Thomas Hubert, Lucas Baker, Matthew Lai, Adrian Bolton,
  et~al.
\newblock Mastering the game of go without human knowledge.
\newblock {\em Nature}, 550(7676):354--359, 2017.

\bibitem[\protect\citeauthoryear{Silver \bgroup \em et al.\egroup
  }{2018}]{silver2018general}
David Silver, Thomas Hubert, Julian Schrittwieser, Ioannis Antonoglou, Matthew
  Lai, Arthur Guez, Marc Lanctot, Laurent Sifre, Dharshan Kumaran, Thore
  Graepel, et~al.
\newblock A general reinforcement learning algorithm that masters chess, shogi,
  and go through self-play.
\newblock {\em Science}, 362(6419):1140--1144, 2018.

\bibitem[\protect\citeauthoryear{Soares \bgroup \em et al.\egroup
  }{2020}]{soares2020explaining}
Eduardo Soares, Plamen~P Angelov, Bruno Costa, Marcos P~Gerardo Castro,
  Subramanya Nageshrao, and Dimitar Filev.
\newblock Explaining deep learning models through rule-based approximation and
  visualization.
\newblock {\em IEEE Transactions on Fuzzy Systems}, 29(8):2399--2407, 2020.

\bibitem[\protect\citeauthoryear{Sundararajan \bgroup \em et al.\egroup
  }{2017}]{sundararajan2017axiomatic}
Mukund Sundararajan, Ankur Taly, and Qiqi Yan.
\newblock Axiomatic attribution for deep networks.
\newblock In {\em Proc. of ICML}, 2017.

\bibitem[\protect\citeauthoryear{Sutton and
  Barto}{2018}]{sutton2018reinforcement}
Richard~S Sutton and Andrew~G Barto.
\newblock {\em Reinforcement learning: An introduction}.
\newblock MIT press, 2018.

\bibitem[\protect\citeauthoryear{Sutton \bgroup \em et al.\egroup
  }{1999}]{sutton1999policy}
Richard~S Sutton, David McAllester, Satinder Singh, and Yishay Mansour.
\newblock Policy gradient methods for reinforcement learning with function
  approximation.
\newblock In {\em Proc. of NeurIPS}, 1999.

\bibitem[\protect\citeauthoryear{Todorov \bgroup \em et al.\egroup
  }{2012}]{todorov2012mujoco}
Emanuel Todorov, Tom Erez, and Yuval Tassa.
\newblock Mujoco: A physics engine for model-based control.
\newblock In {\em Proc. of IROS}, 2012.

\bibitem[\protect\citeauthoryear{Topin \bgroup \em et al.\egroup
  }{2021}]{topin2021iterative}
Nicholay Topin, Stephanie Milani, Fei Fang, and Manuela Veloso.
\newblock Iterative bounding mdps: Learning interpretable policies via
  non-interpretable methods.
\newblock In {\em Proc. of AAAI}, 2021.

\bibitem[\protect\citeauthoryear{Van~Waveren \bgroup \em et al.\egroup
  }{2022}]{van2022correct}
Sanne Van~Waveren, Christian Pek, Jana Tumova, and Iolanda Leite.
\newblock Correct me if i'm wrong: Using non-experts to repair reinforcement
  learning policies.
\newblock In {\em Proc. of HRI}, 2022.

\bibitem[\protect\citeauthoryear{Vinyals \bgroup \em et al.\egroup
  }{2017}]{vinyals2017starcraft}
Oriol Vinyals, Timo Ewalds, Sergey Bartunov, Petko Georgiev, Alexander~Sasha
  Vezhnevets, Michelle Yeo, Alireza Makhzani, Heinrich K{\"u}ttler, John
  Agapiou, Julian Schrittwieser, et~al.
\newblock Starcraft ii: A new challenge for reinforcement learning.
\newblock {\em arXiv preprint arXiv:1708.04782}, 2017.

\bibitem[\protect\citeauthoryear{Vinyals \bgroup \em et al.\egroup
  }{2019}]{vinyals2019grandmaster}
Oriol Vinyals, Igor Babuschkin, Wojciech~M Czarnecki, Micha{\"e}l Mathieu,
  Andrew Dudzik, Junyoung Chung, David~H Choi, Richard Powell, Timo Ewalds,
  Petko Georgiev, et~al.
\newblock Grandmaster level in starcraft ii using multi-agent reinforcement
  learning.
\newblock {\em nature}, 575(7782):350--354, 2019.

\bibitem[\protect\citeauthoryear{Wang \bgroup \em et al.\egroup
  }{2016}]{wang2016dueling}
Ziyu Wang, Tom Schaul, Matteo Hessel, Hado Hasselt, Marc Lanctot, and Nando
  Freitas.
\newblock Dueling network architectures for deep reinforcement learning.
\newblock In {\em Proc. of ICML}, 2016.

\bibitem[\protect\citeauthoryear{Wang \bgroup \em et al.\egroup
  }{2024a}]{wanghelpful}
Jingtan Wang, Xiaoqiang Lin, Rui Qiao, Chuan-Sheng Foo, and Bryan Kian~Hsiang
  Low.
\newblock Helpful or harmful data? fine-tuning-free shapley attribution for
  explaining language model predictions.
\newblock In {\em Proc. of ICML}, 2024.

\bibitem[\protect\citeauthoryear{Wang \bgroup \em et al.\egroup
  }{2024b}]{wang2024rlhfpoison}
Jiongxiao Wang, Junlin Wu, Muhao Chen, Yevgeniy Vorobeychik, and Chaowei Xiao.
\newblock Rlhfpoison: Reward poisoning attack for reinforcement learning with
  human feedback in large language models.
\newblock In {\em Proc. of ACL}, 2024.

\bibitem[\protect\citeauthoryear{Watkins and Dayan}{1992}]{watkins1992q}
Christopher~JCH Watkins and Peter Dayan.
\newblock Q-learning.
\newblock {\em Machine learning}, 8:279--292, 1992.

\bibitem[\protect\citeauthoryear{Yu \bgroup \em et al.\egroup
  }{2023}]{yu2023airs}
Jiahao Yu, Wenbo Guo, Qi~Qin, Gang Wang, Ting Wang, and Xinyu Xing.
\newblock Airs: Explanation for deep reinforcement learning based security
  applications.
\newblock In {\em Proc. of USENIX Security}, 2023.

\bibitem[\protect\citeauthoryear{Yuan \bgroup \em et al.\egroup
  }{2024}]{yuan2024shine}
Zhuowen Yuan, Wenbo Guo, Jinyuan Jia, Bo~Li, and Dawn Song.
\newblock {SHINE}: Shielding backdoors in deep reinforcement learning.
\newblock In {\em Proc. of ICML}, 2024.

\bibitem[\protect\citeauthoryear{Zahavy \bgroup \em et al.\egroup
  }{2016}]{zahavy2016graying}
Tom Zahavy, Nir Ben-Zrihem, and Shie Mannor.
\newblock Graying the black box: Understanding dqns.
\newblock In {\em Proc. of ICML}, 2016.

\end{thebibliography}

\end{document}